\def\year{2019}\relax
\documentclass[letterpaper]{article} 
 \usepackage{booktabs}
\usepackage{aaai19}  
\usepackage{times}  
\usepackage{helvet} 
\usepackage{courier}  
\usepackage[hyphens]{url}  
\usepackage{graphicx} 
\def\UrlFont{\rm}  
\usepackage{graphicx}  
\usepackage{multirow}
\usepackage{acronym}
\usepackage{xspace}
\acrodef{MAPF}{Multi-Agent Pathfinding}
\acrodef{SOC}{Sum-of-Costs}
\newcommand{\mapf}{\ac{MAPF}\xspace}

\usepackage{tikz}
\usetikzlibrary{arrows.meta,shapes}
\usepackage{color}

\usepackage{soul}
\usepackage{float}
\usepackage{graphicx}
\usepackage{xspace}
\usepackage[utf8]{inputenc}
\usepackage[small]{caption}
\usepackage{amsmath}
\usepackage{amsthm}
\usepackage{amssymb}
\usepackage{ifthen}

\newboolean{showcomments}
\setboolean{showcomments}{false}

\ifthenelse{\boolean{showcomments}}
  {\newcommand{\nb}[3]{
  {\color{#2}\small\fbox{\bfseries\sffamily\scriptsize#1}}
  {\color{#2}\sffamily\small$\triangleright~$\textit{\small #3}$~\triangleleft$}
  }
  }
  {\newcommand{\nb}[3]{}
  }

\newcommand{\comment}[1]{{\nb{\textbf{Comment:}}{orange}{#1}}}
\title{Multi-Agent Pathfinding: Definitions, Variants, and Benchmarks}

\author{
Roni Stern,\textsuperscript{\rm 1}
Nathan R. Sturtevant,\textsuperscript{\rm 2}
Ariel Felner,\textsuperscript{\rm 1}
Sven Koenig,\textsuperscript{\rm 3}
Hang Ma,\textsuperscript{\rm 3}\\
\bf \Large Thayne T. Walker,\textsuperscript{\rm 4}
Jiaoyang Li,\textsuperscript{\rm 3}
Dor Atzmon,\textsuperscript{\rm 1}
Liron Cohen,\textsuperscript{\rm 3}
T. K. Satish Kumar,\textsuperscript{\rm 3}
Eli Boyarski,\textsuperscript{\rm 1}
Roman Bart\'{a}k\textsuperscript{\rm 5} \\
\textsuperscript{\rm 1}Ben Gurion University of the Negev,
\textsuperscript{\rm 2}University of Alberta,
\textsuperscript{\rm 3}USC,
\textsuperscript{\rm 4}University of Denver
\textsuperscript{\rm 5}Charles University\\
sternron@post.bgu.ac.il,
nathanst@ualberta.ca,
felner@bgu.ac.il,
skoenig@usc.edu,
hangma@usc.edu,
thayne.walker@du.edu,\\
jiaoyanl@usc.edu,
dorat@post.bgu.ac.il,
lironcoh@usc.edu,
tkskwork@gmail.com,
boyarske@post.bgu.ac.il,
bartak@ktiml.mff.cuni.cz
}

\begin{document}

\maketitle

\begin{abstract}	The \mapf problem is the fundamental problem of planning paths for multiple agents, where the key constraint is that the agents will be able to follow these paths concurrently without colliding with each other. Applications of \mapf include automated warehouses and autonomous vehicles. Research on \mapf has been flourishing in the past couple of years. 
	Different \mapf research papers make different assumptions, e.g., whether agents can traverse the same road at the same time, and have different objective functions, e.g., minimize makespan or sum of agents' actions costs. 
	These assumptions and objectives are sometimes implicitly assumed or described informally. This makes it difficult to establish appropriate baselines for comparison in research papers, as well as making it difficult for practitioners to find the papers relevant to their concrete application. 
	This paper aims to fill this gap and support researchers and practitioners by providing a unifying terminology for describing common \mapf assumptions and objectives. In addition, we also provide pointers to two \mapf benchmarks. In particular, we introduce a new grid-based benchmark for \mapf, and demonstrate experimentally that it poses a challenge to contemporary \mapf algorithms. 
\end{abstract}

\section{Introduction}

\mapf is an important type of multi-agent planning problem in which the task is to plan paths for multiple agents, 
where the key constraint is that the agents will be able to follow these paths concurrently without colliding with each other. \mapf has a range of relevant contemporary applications including automated warehouses, autonomous vehicles, and robotics. Consequently, this problem has received attention in recent years from various research groups and academic communities~\cite{standley2010finding,felner2017search,surynek2016empirical,bartak2018aScheduling,cohen2018anytime,li2019multi,MaAAAI19a}.

Different \mapf research papers consider different sets of assumptions about the agents and aim for different objectives. These assumptions and objectives are sometimes implicitly assumed or described informally. Even in cases where the assumptions and objective function are described formally, there are still differences in used \mapf terminology. This makes it difficult to navigate through and understand existing literature and to establish appropriate baselines for comparison. In addition, it makes it difficult for practitioners to find papers relevant to their concrete application. 

This paper aims to address this growing challenge by introducing  a \textbf{unified terminology} to describe \mapf problems, and by establishing \textbf{common benchmarks and evaluation measures} for evaluating \mapf algorithms. 
The unified \mapf terminology we present in this paper is our attempt to classify the currently studied  \mapf variants. We hope this terminology will serve as a common ground for future researchers, and will be used by them to describe their contributions succinctly and accurately. 

In the second part of this paper, we introduce a new grid \mapf benchmark to the community. This benchmark includes a diverse set of maps, as well as generated source and target vertices. We report the performance of a standard \mapf algorithm on this benchmark, to serve as baseline for comparison to future research. This benchmark is intended to help future researchers and enable more scientifically rigorous empirical comparisons of existing and future \mapf algorithms. We do not claim that these benchmarks are perfect, as they may have some biases. But, through their use and study these biases can be discovered and corrected. It is also important to emphasize that this document is not intended to be a survey of  state of the art MAPF algorithms. For such a survey, see~\cite{felner2017search,ma2017buzz}. 
In addition, the newly created website
\url{http://mapf.info} contains \mapf-related tutorials and other resources. 


\section{Classical MAPF}
We first describe what we refer to as a \emph{classical MAPF} problem. 
The input to a classical MAPF problem with $k$ agents is 
a tuple $\langle G, s, t\rangle$ 
where $G=(V,E)$ is an undirected graph, 
$s:[1,\ldots,k]\rightarrow V$ maps an agent to a source vertex, 
and $t:[1,\ldots,k]\rightarrow V$ maps an agent to a target vertex. 
Time is assumed to be discretized, and in every time step each agent is situated in one of the graph vertices 
and can perform a single \emph{action}. 
An action in classical MAPF is a function $a: V\rightarrow V$ 
such that $a(v)=v'$ means that if an agent is at vertex $v$ and performs $a$ then it will be in vertex $v'$ in the next time step. 
Each agent has two types of actions: \emph{wait} and \emph{move}. 
A \emph{wait} action means that the agent stays in its current vertex another time step. 
A \emph{move} action means that the agent moves from its current vertex $v$ to an adjacent vertex $v'$ in the graph (i.e., $(v,v')\in E$). 

For a sequence of actions $\pi=(a_1,\ldots a_n)$ 
and an agent $i$,  
we denote by $\pi_i[x]$ 
the location of the agent after executing  the first $x$ actions in $\pi$, starting from the agent's source $s(i)$. 
Formally, $\pi_i[x]=a_x(a_{x-1}(\cdots a_1(s(i))))$. 
A sequence of actions $\pi$ is a \textbf{single-agent plan} for agent $i$ iff executing this sequence of actions in $s(i)$ results in being at $t(i)$, 
that is, iff $\pi_i[|\pi|]=t(i)$. 
A \textbf{solution} is a set of $k$ single-agent plans, one for each agent.  

\subsection{Types of Conflicts in Classical MAPF}

The overarching goal of MAPF solvers is to find a solution, i.e., a single-agent plan for each agent, that can be executed without collisions. To achieve this, MAPF solvers use the notion of \emph{conflicts} during planning, where a MAPF solution is called \emph{valid} iff there is no conflict between any two single-agent plans. The definition of what constitutes a conflict depends on the environment, and correspondingly the literature on classical MAPF includes several different definitions of what constitutes a conflict between plans. We list common conflict definitions below. Let $\pi_i$ and $\pi_j$ be a pair of single-agent plans.


\begin{itemize}
\item \textbf{Vertex conflict.} A \emph{vertex conflict} between $\pi_i$ and $\pi_j$ occurs iff  according to these plans the agents are planned to occupy the same vertex at the same time step. 
Formally, there is a vertex conflict between $\pi_i$ and $\pi_j$ iff there exists a time step $x$ such that 
$\pi_i[x]=\pi_j[x]$. 

\item \textbf{Edge conflict.} An \emph{edge conflict} between $\pi_i$ and $\pi_j$ occurs iff 
according to these plans the agents are planned to traverse the same edge at the same time step in the same direction. 
Formally, there is an edge conflict between $\pi_i$ and $\pi_j$ iff there exists a time step $x$ such that 
$\pi_i[x]=\pi_j[x]$ and $\pi_i[x+1]=\pi_j[x+1]$.

\item \textbf{Following conflict.} A \emph{following} conflict between $\pi_i$ and $\pi_j$ occurs iff
one agent is planned to occupy a vertex that was occupied by another agent in the previous time step. 
Formally, there is a following conflict between $\pi_i$ and $\pi_j$ iff there exists a time step $x$ such that 
$\pi_i[x+1]=\pi_j[x]$. 


\item \textbf{Cycle conflict.} A \emph{cycle} conflict  between a set of single-agent plans $\pi_i, \pi_{i+1}, \ldots \pi_j$ occurs iff
in the same time step every agent moves to a vertex that was previously occupied by another agent, forming a ``rotating cycle'' pattern. Formally, a \emph{cycle conflict} between a set of plans $\pi_i, \pi_{i+1}, \ldots \pi_j$ 
occurs iff there exists a time step $x$ in which 
$\pi_i(x+1)=\pi_{i+1}(x)$  
and $\pi_{i+1}(x+1)=\pi_{i+2}(x)$ 
$\ldots$ 
and $\pi_{j-1}(x+1)=\pi_j(x)$ 
and $\pi_j(x+1)=\pi_i(x)$.

\item \textbf{Swapping conflict.} A \emph{swapping} conflict between $\pi_i$ and $\pi_j$ occurs iff 
the agents are planned to swap locations in a single time step. 
Formally, there is a swapping conflict between $\pi_i$ and $\pi_j$ iff there exists a time step $x$ such that 
$\pi_i[x+1]=\pi_j[x]$ and $\pi_j[x+1]=\pi_i[x]$. This conflict is sometimes called \emph{edge conflict} in the current \mapf literature. 

\end{itemize}

\begin{figure}
    \centering
    \includegraphics[width=\columnwidth]{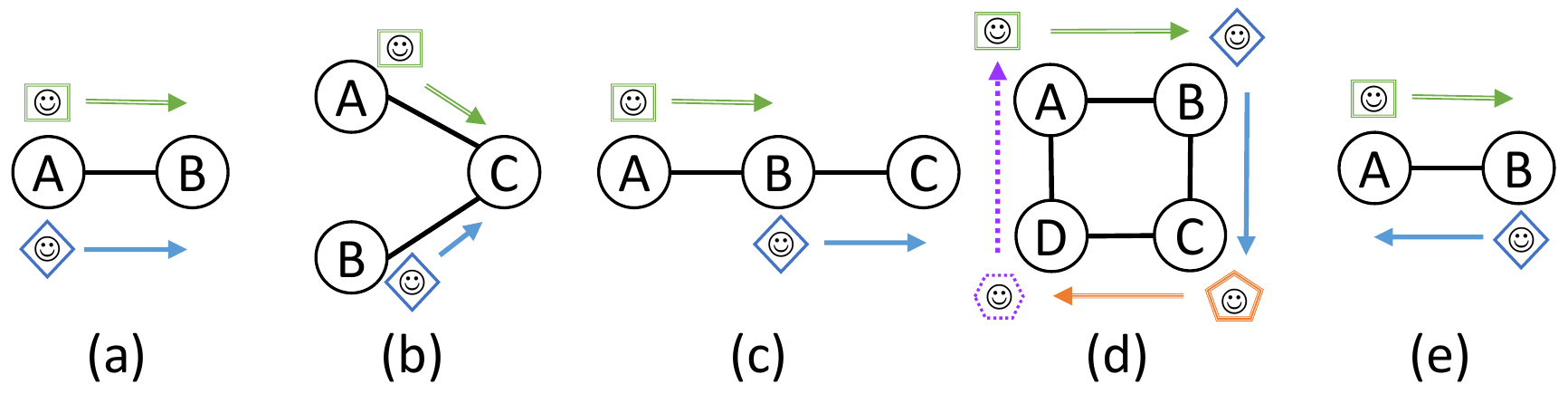}
    \caption{An illustration of common types of conflicts. From left to right: an edge conflict, a vertex conflict, a following conflict, a cycle conflict, and a swapping conflict.}
    \label{fig:types-of-conflicts}
\end{figure}

Figure~\ref{fig:types-of-conflicts} illustrates the different types of conflicts. Note that the above set of conflict definitions is certainly not a complete set of all possible conflicts. 
Considering the formal definitions of these conflicts, it is clear that there are dominance relation between them: (1) forbidding vertex conflicts implies edge conflicts are also forbidden, (2) forbidding following conflicts implies cycle conflicts and swapping conflicts are also forbidden, (3) forbidding cycle conflicts implies that swapping conflicts are also forbidden. Vice versa, (1) allowing edge conflicts implies vertex conflicts are also allowed, 
(2) allowing swapping conflicts implies cycle conflicts are also allowed,\footnote{A swapping conflict is, in fact, a cycle conflict for two agents.}
and (3) allowing cycle conflicts implies following conflicts are also allowed.

To properly define a classical MAPF problem, one needs to specify which types of conflicts are allowed in a solution. 
The least constrained restriction is to only forbid edge conflicts. 
However, to the best of our knowledge, 
all prior work on classical MAPF also forbid vertex conflicts. Some work on MAPF with payload transfers allows swapping conflicts~\cite{MaAAAI16}.
Most work on search-based MAPF algorithms~\cite{standley2010finding,felner2017search} forbid swapping conflicts, but allow following conflicts.
Some work on compilation-based MAPF algorithms 
as well as all work that consider MAPF as a pebble motion problem, forbid following conflicts as well~\cite{surynek2016empirical,bartak2017modeling}. 


\subsection{Agent Behavior at Target in Classical MAPF}

In a solution to a classical MAPF problem, 
agents may reach their targets at different time steps.
Therefore, when defining a classical MAPF problem one must define 
how an agent behaves in the time steps after it has reached its target and before the last agent has reached its target. 

There are two common assumptions for how agents behave at their targets.
\begin{itemize}
\item \textbf{Stay at target} Under this assumption, 
an agent waits in its target until all agents have reached their targets. 
This waiting agent will cause a vertex conflict with any plan that passes through
its target after it has reached it. 
Formally, under the stay-at-target assumption, 
a pair of single-agent plans $\pi_i$ and $\pi_j$ will have a vertex conflict 
if there exists a time step $t\geq |\pi_i|$ such that $\pi_j[t]=\pi_i[|\pi_i|]$.
\item \textbf{Disappear at target} Under this assumption, 
when an agent reaches its target it immediately disappears. 
This means the plan of that agent will not have any conflict after 
the time step in which the corresponding agent has reached its target. 
\end{itemize}
Most prior work on classical MAPF assumed stay-at-target, but recent work also considered the disappear-at-target assumption~\cite{MaAAAI19a}.

\subsection{Objective Functions in Classical MAPF}
It is safe to say that in most real applications of MAPF, 
some MAPF solutions are better than others. 
To capture that, work in classical MAPF considers an \textbf{objective function}
that is used to evaluate MAPF solutions. 
The two most common functions used for evaluating a solution in classical MAPF are \emph{makespan} and \emph{sum of costs}. 
\begin{itemize}
\item \textbf{Makespan.} The number of time steps required for all agents to reach their target. 
For a MAPF solution $\pi=\{\pi_1, \ldots \pi_k\}$, the makespan of $\pi$ is defined as $\max_{1\leq i\leq k} |\pi_i|$. 
\item \textbf{Sum of costs.} The sum of time steps required by each agent to reach its target. The sum of costs of $\pi$ is defined as $\sum_{1\leq i\leq k} |\pi_i|$. 
Sum of costs is also known as \emph{flowtime}.
\end{itemize}


If the agent-at-target behavior is \emph{stay at target} and the objective function is \emph{sum of costs}, then one needs to specify how  staying at a target affects the sum of costs. For example, one can define that, if an agent waits at its target, then it does not increase the sum of costs. The common assumption in most prior work is that an agent staying in its target counts as a wait action unless it is not planning to move away from its target again. For example, assume that agent $i$ reaches its target at time step $t$, leaves its target at time step $t'$, arrives back at its target at time step $t''$, 
and then stays at its target until all agents reach their target. 
Then, this single-agent plan will contribute $t''$ to the sum of costs of the corresponding solution. 

We do not claim that these are the only possible objective functions for classical MAPF. 
One may define other objective functions, such as the total non-waiting actions required to reach the target (some refer to this as the sum-of-fuel), and total time spent by the agent not in the target. 
However, to the best of our knowledge, the above objective functions are the only ones used in prior work on classical MAPF. Makespan has been used extensively by compilation-based MAPF algorithms, while sum of costs has been used by most search-based MAPF algorithms. But, there has also been work on both objective functions by both types of MAPF algorithms~\cite{surynek2016empirical}. There has also been work on maximizing the number of agents reaching their targets within a given makespan (i.e., deadline)~\cite{MaIJCAI18}.


\section{Beyond Classical MAPF}
All the above classical MAPF variants share the following assumptions: (1) time is discretized into time steps, (2) every action takes exactly one time step, 
and (3) in every time step, each agent occupies exactly a single vertex. 

Next, we briefly list several MAPF variants that relax these assumptions.

\subsection{MAPF on Weighted Graphs}
The assumption that each action -- move or wait -- takes exactly one time step, implicitly assumes a somewhat simplistic motion model for the agents. 
More complex motion models have been studied in the MAPF literature, in which different actions may have different duration. 
This means the underlying graph that represents the possible locations agents may occupy (denoted $G$ earlier) is now a weighted graph, where the weight of each edge represents the duration it will take an agent to traverse this edge.\footnote{One can also differentiate between the time it takes to traverse an edge and the cost it incurs. E.g., it may take one time step to traverse an edge but it may cost more, for example, energy.}
\begin{figure}

\label{next}
\begin{center}
\setlength\tabcolsep{1.5pt}
\begin{tabular}{cccc}

\begin{tikzpicture}
\draw[step=0.4cm,color=gray] (-3.6,-.4) grid (-2.4,0.8);
\draw[->,>=stealth] (-3,.2) -- (-2.6,0.2);
\draw[->,>=stealth] (-3,.2) -- (-3,0.6);
\draw[->,>=stealth] (-3,.2) -- (-3,-0.2);
\draw[->,>=stealth] (-3,.2) -- (-3.4,0.2);
\end{tikzpicture}
&
\begin{tikzpicture}
\draw[step=0.4cm,color=gray] (-.4,-.4) grid (.8,.8);
\draw[->,>=stealth] (.2,.2) -- (.6,0.2);
\draw[->,>=stealth] (.2,.2) -- (.6,0.6);
\draw[->,>=stealth] (.2,.2) -- (.6,-0.2);
\draw[->,>=stealth] (.2,.2) -- (.2,0.6);
\draw[->,>=stealth] (.2,.2) -- (.2,-0.2);
\draw[->,>=stealth] (.2,.2) -- (-.2,0.2);
\draw[->,>=stealth] (.2,.2) -- (-.2,0.6);
\draw[->,>=stealth] (.2,.2) -- (-.2,-0.2);
\end{tikzpicture}
&
\begin{tikzpicture}
\draw[step=0.4cm,color=gray] (-.8,-.8) grid (1.2,1.2);
\draw[->,>=stealth] (.2,.2) -- (1,0.6);
\draw[->,>=stealth] (.2,.2) -- (1,-0.2);
\draw[->,>=stealth] (.2,.2) -- (.6,0.2);
\draw[->,>=stealth] (.2,.2) -- (.6,0.6);
\draw[->,>=stealth] (.2,.2) -- (.6,1);
\draw[->,>=stealth] (.2,.2) -- (.6,-0.2);
\draw[->,>=stealth] (.2,.2) -- (.6,-0.6);
\draw[->,>=stealth] (.2,.2) -- (.2,0.6);
\draw[->,>=stealth] (.2,.2) -- (.2,-0.2);
\draw[->,>=stealth] (.2,.2) -- (-.2,0.2);
\draw[->,>=stealth] (.2,.2) -- (-.2,0.6);
\draw[->,>=stealth] (.2,.2) -- (-.2,1);
\draw[->,>=stealth] (.2,.2) -- (-.2,-0.2);
\draw[->,>=stealth] (.2,.2) -- (-.2,-0.6);
\draw[->,>=stealth] (.2,.2) -- (-.6,0.6);
\draw[->,>=stealth] (.2,.2) -- (-.6,-0.2);
\end{tikzpicture}
&
\begin{tikzpicture}
\draw[step=0.4cm,color=gray] (-4.4,-1.2) grid (-1.6,1.6);
\draw[->,>=stealth] (-3,.2) -- (-1.8,0.6);
\draw[->,>=stealth] (-3,.2) -- (-1.8,1);
\draw[->,>=stealth] (-3,.2) -- (-1.8,-0.2);
\draw[->,>=stealth] (-3,.2) -- (-1.8,-0.6);
\draw[->,>=stealth] (-3,.2) -- (-2.2,1.4);
\draw[->,>=stealth] (-3,.2) -- (-2.2,0.6);
\draw[->,>=stealth] (-3,.2) -- (-2.2,-0.2);
\draw[->,>=stealth] (-3,.2) -- (-2.2,-1);
\draw[->,>=stealth] (-3,.2) -- (-2.6,1.4);
\draw[->,>=stealth] (-3,.2) -- (-2.6,0.2);
\draw[->,>=stealth] (-3,.2) -- (-2.6,0.6);
\draw[->,>=stealth] (-3,.2) -- (-2.6,1);
\draw[->,>=stealth] (-3,.2) -- (-2.6,-0.2);
\draw[->,>=stealth] (-3,.2) -- (-2.6,-0.6);
\draw[->,>=stealth] (-3,.2) -- (-2.6,-1);
\draw[->,>=stealth] (-3,.2) -- (-3,0.6);
\draw[->,>=stealth] (-3,.2) -- (-3,-0.2);
\draw[->,>=stealth] (-3,.2) -- (-3.4,1.4);
\draw[->,>=stealth] (-3,.2) -- (-3.4,0.2);
\draw[->,>=stealth] (-3,.2) -- (-3.4,0.6);
\draw[->,>=stealth] (-3,.2) -- (-3.4,1);
\draw[->,>=stealth] (-3,.2) -- (-3.4,-0.2);
\draw[->,>=stealth] (-3,.2) -- (-3.4,-0.6);
\draw[->,>=stealth] (-3,.2) -- (-3.4,-1);
\draw[->,>=stealth] (-3,.2) -- (-3.8,1.4);
\draw[->,>=stealth] (-3,.2) -- (-3.8,0.6);
\draw[->,>=stealth] (-3,.2) -- (-3.8,-0.2);
\draw[->,>=stealth] (-3,.2) -- (-3.8,-1);
\draw[->,>=stealth] (-3,.2) -- (-4.2,0.6);
\draw[->,>=stealth] (-3,.2) -- (-4.2,1);
\draw[->,>=stealth] (-3,.2) -- (-4.2,-0.2);
\draw[->,>=stealth] (-3,.2) -- (-4.2,-0.6);
\end{tikzpicture}
\end{tabular}
\end{center}
\caption{$2^k$ Neighborhood movement models for k = 2,3,4 and 5.}
\label{fig:k-neighborhood}
\end{figure}
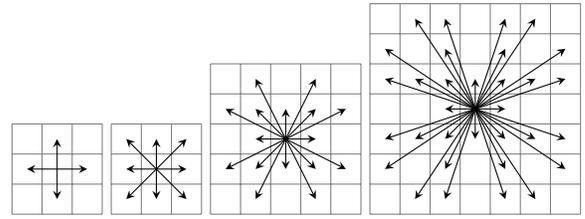


Bartak et al.~\shortcite{bartak2018aScheduling} proposed a scheduling-based approach for \mapf on weighted graphs, and Walker et al.~\shortcite{walker2018extended} proposed a variant of the Increasing Cost Tree Search (ICTS) algorithm. Yakovlev and Andreychuk~\shortcite{yakovlev2017any} proposed a hybrid of the SIPP algorithm~\cite{phillips2011sipp} and prioritized planning for weighted graphs. 

The types of weighted graphs that have been used in MAPF reseach so far include: 
\begin{itemize}
    \item \textbf{MAPF in $2^k$-neighbor grids.}\footnote{Such grids are also referred to as $2^k$-connected grids.} Such maps are a restricted form of weighted graphs in which every vertex represents a cell in a two-dimensional grid. The move actions of an agent in a cell are all its $2^k$ neighboring cells, where $k$ is a parameter.
Costs are based on Euclidean distance, therefore when $k>2$, this introduces actions with different costs. For example, in an 8-neighbor grid a diagonal move costs $\sqrt{2}$ while a move in one of the cardinal directions costs 1. Figure~\ref{fig:k-neighborhood} shows the possible move actions in $2^k$-neighbor grids for $k=2, 3, 4,$ and $5$. 


    \item \textbf{MAPF in Euclidean space.} 
\mapf in Euclidean space is a generalization of \mapf in which every node in $G$ represents a Euclidean point ($x,y$), and the edges represent allowed move actions. Such settings arise, for example, when the underlying graph is a roadmap generated for a continuous Euclidean environment \cite{khatib1986real,wagner2012probabilistic}. 

\end{itemize} 


\subsection{Feasibility Rules}

The definition of a valid solution used in classical MAPF -- no conflicts -- is just one 
type of solution requirement. 
We use the term \emph{feasibility rule} to refer to a requirement over a \mapf solution. Other \mapf feasibility rules have been suggested.

\begin{itemize}
\item \textbf{Robustness rules.} These are rules designed to ensure that a \mapf solution considers inadvertent delays in execution. A \emph{k-robust} MAPF plan builds in a sufficient buffer for agents to be delayed up to $k$ time steps without resulting in a conflict~\cite{atzmon2018robust}. 
When the probability of future delays is known, robustness rules can require that the probability an agent will conflict during execution is lower than a given bound~\cite{wagner2017path} or be combined with execution policies to guarantee a conflict-free execution \cite{MaAAAI17}.

\item \textbf{Formation rules.} These are restrictions over the allowed move actions of an agent that depend on the location of the other agents but are not related to collisions. For example, restrictions intended for the agents to maintain a specified formation~\cite{barel2017come}, or to maintain a communication link with a set of neighboring agents~\cite{stump2011visibility,gilboa2006distributed}. 

\end{itemize}



\subsection{From Pathfinding to Motion Planning}
In classical MAPF, agents are assumed to occupy exactly one vertex, in a sense having no volume, no shape, and move at constant speed. By contrast, motion planning algorithms directly consider these  properties. There, an agent is situated at each time step in a \emph{configuration} instead of only a vertex, where a configuration specifies the agent's location, orientation, velocity, etc, and an edge between configurations represents kinematic motion. Several notable \mapf variants are steps towards closing this gap between classical \mapf and motion planning. 

\subsubsection{\mapf with large agents.}
Some \mapf research considered agents with a specific geometric shape and volume~\cite{li2019multi,walker2018extended,yakovlev2017any,thomas2015extended}. The fact that agents have volume raises questions about how they are situated in the underlying graph $G$ and how they move in it. In particular, if an agent is located in one vertex, it may prohibit other agents from occupying nearby vertices. 
Similarly, if an agent moves along an edge it may prohibit other agents from moving along intersecting edges or staying at vertices that are too close to the edge. This may introduce new types of conflicts, such as vertex-to-vertex, edge-to-edge, and edge-to-vertex conflicts \cite{honig2018trajectory}. 

Several approaches for solving \mapf with large agents have appeared in the literature, including a CBS-based approach~\cite{li2019multi}, an ICTS-based approach~\cite{walker2018extended}, and a prioritized planning approach~\cite{yakovlev2017any}. A special case of agents with volume is the \emph{convoy} setting, in which agents occupy a string of vertices and their connecting edges \cite{thomas2015extended}.

\subsubsection{\mapf with kinematic constraints.} 
Other \mapf research considered kinematic constraints over agents' move actions~\cite{honig2017summary,walker17hierarchical}. That is, the move actions an agent can perform depend not only on its current location, but also on state parameters such as velocity and orientation. A by-product of such constraints is that the underlying graph becomes directed, as there may be edges that can only be passable in one direction due to kinematic constraints of the agent.  MAPF-POST, as an example, is a \mapf algorithm that considers these kinematic constraints by post-processing a solution created by a \mapf algorithm. There is also a reduction-based approach that assumes rotation actions as a half way to kinematic constraints~\cite{BartakIBERAMIA18}.

\subsection{Tasks and Agents}

In classical MAPF, each agent has one task - to get it to its target. Several extensions have been made in the MAPF literature in which agents may be assigned more than one target.  

\subsubsection{Anonymous \mapf.} 
In this \mapf variant, the objective is to move the agents to a set of target vertices, but it does not matter which agent reaches which target~\cite{kloder2006path,yu2013multi}. 
Another way to view this \mapf variant is as a \mapf problem in which every agent can be assigned to any target, but it has to be a one-to-one mapping between agents and targets. 

\subsubsection{Colored \mapf.}
This \mapf variant is a generalization of anonymous \mapf in which agents are grouped into teams, and every team has a set of targets. The objective is to move the agents in each team to their targets~\cite{ma2016optimal,solovey2014k}. Another way to view this \mapf variant is as a \mapf problem in which every agent can be assigned to targets only from the set of targets designated for its team. 

One can generalize colored \mapf even further, assigning a target and an agent to multiple teams. 

\subsubsection{Online MAPF.}
In \emph{online \mapf}, a sequence of \mapf problems are solved on the same graph. This setting has also been called ``Lifelong MAPF''~\cite{MaAAMAS17,MaAAAI19b}. Online \mapf problems can be classified as follows.
\begin{itemize}
    \item \textbf{Warehouse model.} This is the setting where a fixed set of agents solve a \mapf problem, but after an agent finds a target, it may be tasked to go to a different target~\cite{MaAAAI19b}. This setting is inspired by \mapf for autonomous warehouses.
    \item \textbf{Intersection model.} This is the setting where new agents may appear, and each agent has one task -- to reach its target~\cite{svancara2019online}. This setting is inspired by autonomous vehicles entering and exiting intersections~\cite{dresner2008multiagent}. 
\end{itemize}
Of course, hybrid models in which an agent can receive a new task when it reaches its target and new agents can appear over time is also possible. 

\section{Benchmarks}

In this section, we describe how classical \mapf algorithms have been evaluated, suggest an organized benchmark for this purpose, and point to other relevant benchmark suites. 

\subsection{Characteristics of a \mapf Benchmark}
A \mapf problem is defined by a graph and a set of source and target vertices. As such, a benchmark for \mapf includes a set of graphs, and for every graph a set of sets of source and target vertices. 
\subsubsection{Graphs for evaluating \mapf algorithms.}
\begin{figure}[tb]
    \centering
    \includegraphics[width=\columnwidth]{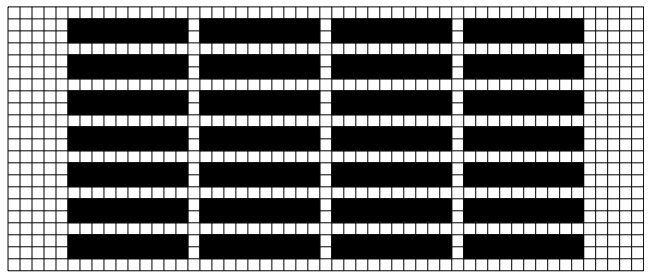}
    \caption{Illustration of a warehouse grid~\cite{CWKKCS:AAMAS:18}.}
    \label{fig:warehouse}
\end{figure}

The types of maps commonly used in prior work include:
\begin{itemize}
    \item \textbf{Dragon Age Origins (DAO) maps.} These are grids taken from the game Dragon Age Origin and are publicly available in Sturtevant's \texttt{movingai.com}  repository~\cite{sturtevant2012benchmarks}. 
    These grids are relatively large and open, where some grids are as large as a $1000\times 1000$ and more. 
    
    \item \textbf{Open $N\times N$ grids.} These are $N\times N$ grids, where common values of $N$ are 8, 16, and 32. Such grids allow experiments in which the ratio of agents to space or \emph{agent density} is high, having fewer vertices without an agent in them. 
    
    \item \textbf{$N\times N$ grids with random obstacles.} These are  $N\times N$ grids, where a set of grid cells are randomly  selected and are considered to be impassable (obstacles)~\cite{standley2010finding}. 
    
    \item \textbf{Warehouse grids.} Inspired by real-world autonomous warehouse applications, recent \mapf papers also experimented with grids shaped to be similar to an automated warehouse, with long corridors~\cite{MaAAMAS17,cohen2018anytime}. Figure~\ref{fig:warehouse} shows an illustration of a warehouse grid taken from~\cite{cohen2018anytime}.
\end{itemize}

\comment{Roni: I wonder if we should add roadmaps. Thoughts?}

\subsection{Sources and targets assignments}
\label{sec:source-target-assignment}

After choosing a type of map, one needs to set the agents' source and target vertices. 
Several methods for setting agents' sources and targets have been used in the literature, including:

\begin{itemize}
    \item \textbf{Random.} Setting the source and target vertices by randomly choosing vertices and making sure there is a path in the graph between them. 
    \item \textbf{Clustered.} Setting the first agent's source and target by randomly choosing vertices in the graph. Setting the sources and targets of all others agents to be with distance of at most $r$ from the first agent's source and target, respectively, where $r$ is a parameter. 
    \item \textbf{Designated.} Setting the source of each agent by randomly choosing from a designated set of possible source vertices, and setting the target of each agent similarly by choosing from a set designated set of possible target vertices. 
\end{itemize}

\emph{Random} assignment is probably the most common in the literature. The \emph{clustered} assignment method has been used to make \mapf problems more challenging.\comment{Roni: TODO: Add a reference of someone doing this}
The \emph{designated} assignment method has been used in prior work in an effort to simulate automated warehouses~\cite{DBLP:conf/socs/CohenUK15,MaAAAI19a,MaAAAI19b} and autonomous vehicles in intersections~\cite{svancara2019online}. In an automated warehouse, there are often humans situated in specific locations to package the delivered bin and most tasks deliver a package to, or retrieve a package from, a human in these locations. In a setting with autonomous  vehicles driving in and out of an intersection, the designated sources and targets are the intersection end points~\cite{svancara2019online}.

\begin{figure}[tb]
    \centering
    \includegraphics[width=\columnwidth]{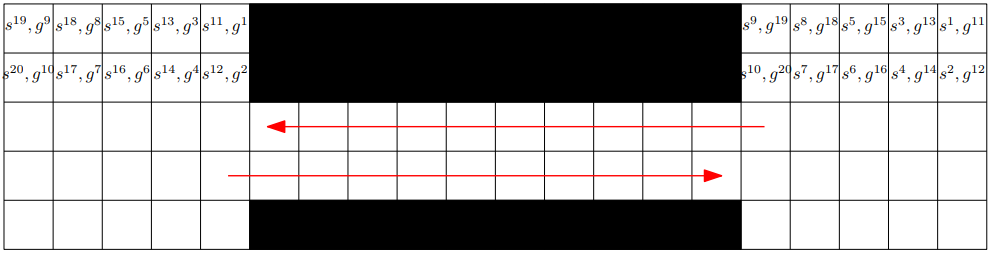}
    \caption{An example of a designated method for setting sources and targets, taken from~\cite{DBLP:conf/socs/CohenUK15}. They chose randomly a source from the open space on the left and a target from the open space on the right for 25\% of the agents, and chose sources and targets from open spaces on the right and left, respectively, for the rest of the agents.}
    \label{fig:designated}
\end{figure}

\subsection{Publicly Available \mapf Benchmarks}

We describe here two publicly available benchmarks for \mapf research, the first of which is a new benchmark set described for the first time in this paper.

\subsubsection{Grid-based \mapf.}

\begin{figure}[tb]
    \centering
    \includegraphics[width=\columnwidth]{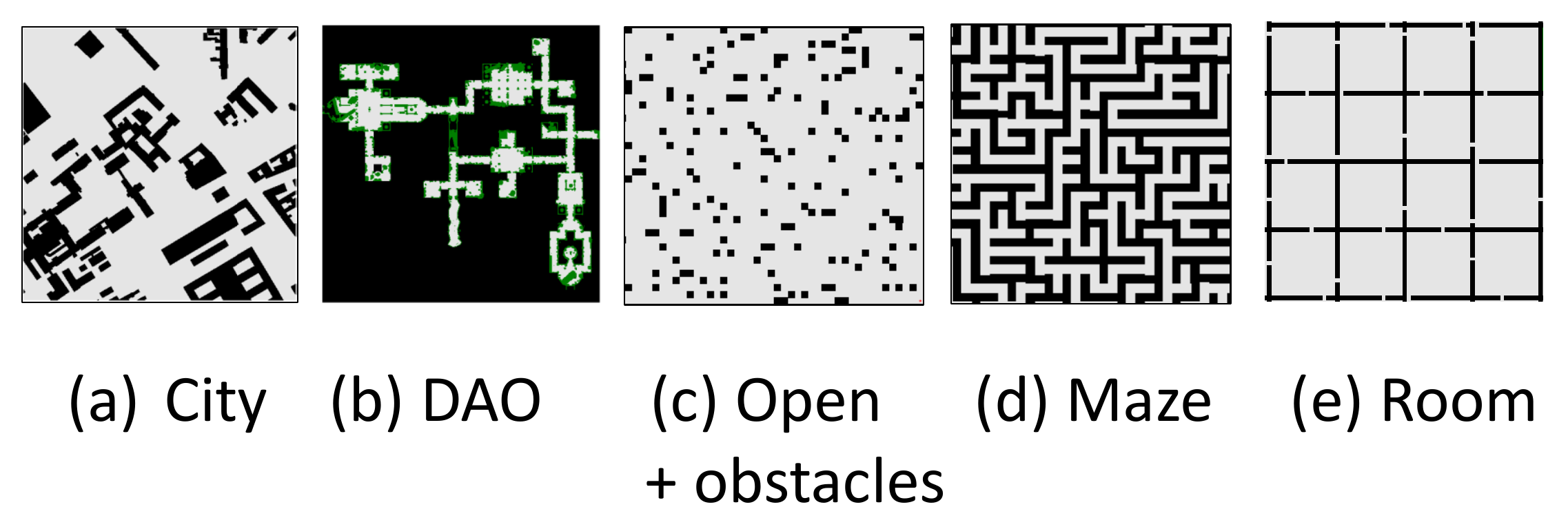}
    \caption{An example of a map from each type of maps available in the grid \mapf benchmark.}
    \label{fig:map-types}
\end{figure}

\begin{table}
\resizebox{\columnwidth}{!}{
\begin{tabular}{@{}llcrrrr@{}}
\toprule
\multicolumn{1}{l}{Type}                                                   & \multicolumn{1}{l}{Map} & Size       & \multicolumn{1}{c}{Problems} & \multicolumn{1}{c}{Solved} & \multicolumn{1}{c}{Min} & \multicolumn{1}{c}{Max} \\ \midrule
\multirow{3}{*}{City}                                                      & Berlin\_1\_256          & 256 X 256  & 25000    & 892    & 4                       & 52                      \\
                                                                           & Boston\_0\_256          & 256 X 256  & 25000    & 718    & 13                      & 47                      \\
                                                                           & Paris\_1\_256           & 256 X 256  & 25000    & 805    & 3                      & 47                      \\
\midrule \multirow{6}{*}{DAO}                                                       & brc202d                 & 481 X 530  & 25000    & 252    & 2                       & 22                      \\
                                                                           & den312d                 & 81 X 65    & 25000    & 577    & 7                      & 36                      \\
                                                                           & den520d                 & 257 X 256  & 25000    & 661    & 5                      & 47                      \\
                                                                           & lak303d                 & 194 X 194  & 25000    & 377    & 8                      & 27                      \\
                                                                           & orz900d                 & 656 X 1491 & 25000    & 162    & 2                       & 12                       \\
                                                                           & ost003d                 & 194 X 194  & 25000    & 535    & 7                      & 37                      \\
\midrule \multirow{4}{*}{\begin{tabular}[c]{@{}l@{}}Dragon \\ Age 2\end{tabular}}   & ht\_chantry             & 141 X 162  & 25000    & 513    & 11                      & 35                      \\
                                                                           & ht\_mansion\_n          & 270 X 133  & 25000    & 795    & 17                      & 42                      \\
                                                                           & w\_woundedcoast         & 578 X 642  & 25000    & 336    & 6                       & 22                      \\
                                                                           & lt\_gallowstemplar\_n   & 180 X 251  & 25000    & 493    & 10                       & 32                      \\
\midrule \multirow{4}{*}{Open}                                                      &                                                                             empty-8-8               & 8 X 8      & 800      & 528    & 18                      & 25                      \\
& empty-16-16             & 16 X 16    & 3200     & 840    & 15                       & 52                      \\
                                                                           & empty-32-32             & 32 X 32    & 12800    & 1190   & 12                       & 81                      \\
                                                                           & empty-48-48             & 48 X 48    & 25000    & 1349   & 18                      & 118                     \\

\midrule \multirow{4}{*}{\begin{tabular}[c]{@{}l@{}}Open+\\ obstacles\end{tabular}} & random-32-32-10         & 32 X 32    & 11525    & 1027   & 15                      & 68                      \\
                                                                           & random-32-32-20         & 32 X 32    & 10225    & 862    & 15                      & 46                      \\
                                                                           & random-64-64-10         & 64 X 64    & 25000    & 1450   & 24                      & 87                      \\
                                                                           & random-64-64-20         & 64 X 64    & 25000    & 1078   & 10                      & 64                      \\
\midrule \multirow{4}{*}{Maze}                                                                           & maze-32-32-2            & 32 X 32    & 8325     & 327    & 8                       & 17                      \\
                                                                           & maze-32-32-4            & 32 X 32    & 9875     & 317    & 4                       & 23                      \\
                                                                           & maze-128-128-10         & 128 X 128  & 25000    & 272    & 5                       & 20                      \\
                                                                           & maze-128-128-2          & 128 X 128  & 25000    & 178    & 4                       & 15                      \\

\midrule \multirow{3}{*}{Room}                                                      & room-32-32-4            & 32 X 32    & 12800    & 469    & 10                      & 26                      \\
                                                                           & room-64-64-16           & 64 X 64    & 25000    & 629    & 12                      & 45                      \\
                                                                           & room-64-64-8            & 64 X 64    & 25000    & 360    & 8                       & 24                      \\ \bottomrule
\end{tabular}
}
\caption{Results for running ICBS with a timeout of 30 seconds on the grid \mapf benchmark.}
\label{tab:results}
\end{table}

This publicly available\footnote{\url{https://movingai.com/benchmarks/mapf.html}} benchmark consists of 24 maps taken from (1) maps of real cities, (2) the video games Dragon Age Origins and Dragon Age 2, (3) open grids with and without random obstacles, (4) maze-like grids, and (5) room-like grids. All maps were taken from the MovingAI pathfinding repository~\cite{sturtevant2012benchmarks}.\footnote{\url{https://movingai.com/benchmarks/grids.html}} 
Figure~\ref{fig:map-types} shows an example of a map from each of these types, 
and Table~\ref{tab:results} shows the dimensions of these maps.

Every map has 25 \emph{scenarios}. 
Every scenario has a list of source and target vertices that 
were set using a variant of the random method (see Section~\ref{sec:source-target-assignment}) All points in the largest reachable region of each map were randomly paired, and then the first 1000 problems were put into the scenario. Thus, one can create a set of \mapf problems from each scenario by choosing any subset of source and target vertices. 

We propose using this benchmark in the following way. 
For a chosen \mapf algorithm, map type, and scenario, 
try to solve as many agents as possible in each scenario, adding them in consecutive order. That is, start by creating a \mapf problem of two agents, using the first two source-target pairs associated with the chosen scenario, 
and run the \mapf algorithm of choice to solve this problem. 
If the algorithm of choice successfully solves this \mapf problem in reasonable time, create a new \mapf problem with 3 agents by using the first three source-target pairs of that scenario and try to solve it with the \mapf algorithm of choice. This continues iteratively until the algorithm of choice cannot solve the created \mapf problem in reasonable time. An evaluated algorithm can then report, for every scenario, the maximal number of agents it was able to solve in reasonable time. 

\comment{Nathan: Please use the terminology from this paper to describe the ICBS settings used for this experiment.}
To provide a baseline for comparison, we performed this evaluation process using ICBS~\cite{boyarski2015icbs}. Using the terminology introduced in this paper, our setting was a classical \mapf setting on a 4-neighbor grid 
where 
(1) edge, vertex, and swapping conflicts were forbidden, 
(2) following and cycle conflicts were allowed, 
(3) the objective was the sum of costs, and 
(4) the agent behavior at target is stay at target. 

Table~\ref{tab:results} shows the result of this evaluation. We set 30 seconds as the runtime limit. 
The different rows correspond to different maps. 
The ``Size'' column reports the number of rows and columns in each map. 
The ``Problems'' column reports the number of problems available for each map.
Note that this number is aggregated over the 25 scenarios, where the number of problems available in a scenario is the number of source-target pairs defined for it. \comment{Roni: is the above clear? Nathan: Mostly - I don't see a good way to improve it.}
The ``Solved'' column reports the number of problems solved by ICBS under the specified time limit (30 seconds). As can be seen, while ICBS is able to solve many problems, the problems in this benchmark are complex enough so that there are many problems that cannot be solved by ICBS in the allotted time. Thus, the problems in this grid \mapf benchmark are hard enough to pose a challenge for contemporary \mapf solvers.

Table~\ref{tab:results} has two additional columns - ``Min'' and ``Max''. These columns report the maximal number of agents solved by ICBS 
in the scenario in which this number was smallest (``Min'') and when it was largest (``Max''). For example, the ``Min'' value for map \texttt{brc202d} is 2 and the ``Max'' is  22. This means there exists a scenario for this map in which ICBS were able to solve at most 2 agents before reaching a timeout, 
and there is a different scenario for this map in which ICBS were able to solve up to 22 agents. We report these values to show the diversity in difficulty between the scenarios of the same map.  

\subsubsection{Asprilo.}

\begin{figure}[tb]
    \centering
    \includegraphics[width=\columnwidth]{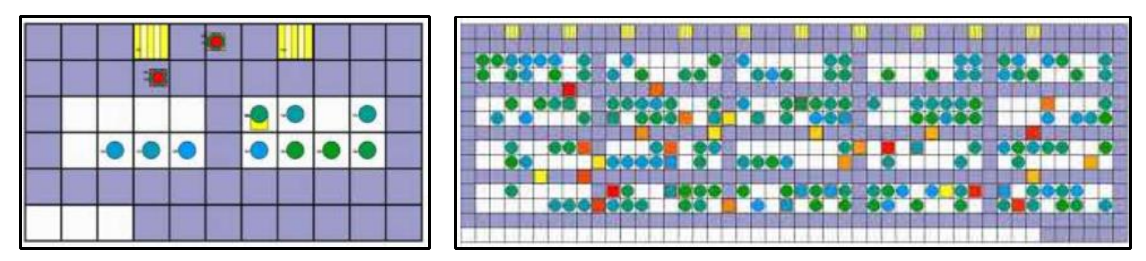}
    \caption{Two scenarios from the Asprilo framework. The left figure is a full warehouse scenario, which includes moving bins from one place to another. The right figure is a movement-only scenario, which corresponds to a classical \mapf problem.}
    \label{fig:asprilo}
\end{figure}

An additional tool that is useful for \mapf research is Asprilo.
Asprilo is a publicly available framework for simulating an automated warehouse~\cite{martin2018experimenting}. It includes tools for defining and generating standard automated warehouse planning problems, and tools for verifying and visualizing plans that solve these problems. 

The type of planning problems supported by Asprilo includes problems in which robots are tasked to pick up and deliver bins in the warehouse from one place to another. These scenarios are grouped into different \emph{domains}, representing different types of problems. Of interest to the \mapf community in particular is domain $M$, which basically represents \mapf problems. Thus, one can use problems from this domain as a benchmark for \mapf algorithms. 
Figure~\ref{fig:asprilo} shows two scenarios from Asprilo. The scenario depicted on the left side is a full warehouse scenario, where the agents are tasked to move bins from one place to another. The scenario on the right side is a movement-only scenario, i.e., a classical \mapf problem. 
Details on ASPRILO can be found in~\cite{martin2018experimenting}, as well as the project's website.\footnote{\url{https://asprilo.github.io}} 

\section{Conclusion}
In the first part of this paper, we defined common assumptions in the ``classical'' Multi-Agent Pathfinding (\mapf) problem and discuss the relationships between them. Then, we defined notable extensions to classical \mapf that were previously published. 
In the second part of this paper, we introduced a new suite of \mapf benchmark problems and point to another set of \mapf benchmark problems. 
Both parts of this paper are intended to propose a common language, terminology, and experimental setting for \mapf research. It is our hope that future \mapf researchers will follow our terminology and find these benchmarks useful. 

\section{Acknowledgments}
This research is supported by ISF grants no. 210/17 to Roni Stern and \#844/17 to Ariel Felner and Eyal Shimony, by BSF grant \#2017692 and by NSF grant \#1815660. 

\vskip 0.2in
\bibliography{library}
\bibliographystyle{aaai}

\end{document}